\documentclass[runningheads]{llncs}

\usepackage{eccv}

\usepackage{wrapfig}

\usepackage{eccvabbrv}

\usepackage{graphicx}
\usepackage{booktabs}

\usepackage{tabularx}
\usepackage{multirow}
\usepackage{textcomp}
\usepackage{adjustbox}
\usepackage{tabularx,booktabs,makecell,array}
\usepackage{float}
\definecolor{mypink}{RGB}{251,49,153}
\usepackage[accsupp]{axessibility}  

\usepackage{hyperref}

\usepackage{orcidlink}

\begin{document}

\title{Language-Guided Token Compression with Reinforcement Learning in Large Vision-Language Models}

\titlerunning{TPRL}

\author{Sihan Cao\inst{1} \and
Jianwei Zhang\inst{1} \and
Pengcheng Zheng\inst{1}
\and
Jiaxin Yan\inst{1}
\and
Caiyan Qin\inst{2}
\and
Yalan Ye\inst{1}
\and
Wei Dong\inst{3}
\and
Peng Wang\inst{1}
\and
Yang Yang\inst{1}
\and
Chaoning Zhang\inst{1}
}

\authorrunning{S.~Cao et al.}

\institute{School of Computer Science and Engineering,
University of Electronic Science and Technology of China, Chengdu, China \and
School of Robotics and Advanced
Manufacture, Harbin Institute of
Technology, Shenzhen, China \and
College of Information and Control Engineering, Xi'an University of Architecture and Technology, Xi'an, China \\
\color{mypink}{\url{https://github.com/MagicVicCoder/TPRL}} 
}

\maketitle

\begin{abstract}
Large Vision-Language Models (LVLMs) incur substantial inference costs due to the processing of a vast number of visual tokens. Existing methods typically struggle to model progressive visual token reduction as a multi-step decision process with sequential dependencies and often rely on hand-engineered scoring rules that lack adaptive optimization for complex reasoning trajectories. To overcome these limitations, we propose TPRL, a reinforcement learning framework that learns adaptive pruning trajectories through language-guided sequential optimization tied directly to end-task performance. We formulate visual token pruning as a sequential decision process with explicit state transitions and employ a self-supervised autoencoder to compress visual tokens into a compact state representation for efficient policy learning. The pruning policy is initialized through learning from demonstrations and subsequently fine-tuned using Proximal Policy Optimization (PPO) to jointly optimize task accuracy and computational efficiency. Our experimental results demonstrate that TPRL removes up to 66.7\% of visual tokens and achieves up to a 54.2\% reduction in FLOPs during inference while maintaining a near-lossless average accuracy drop of only 0.7\%. Code is released at \href{https://github.com/MagicVicCoder/TPRL}{\textcolor{mypink}{https://github.com/MagicVicCoder/TPRL}}.
\end{abstract}

\section{Introduction}
\label{sec:intro}

Large Vision-Language Models (LVLMs)~\cite{bai2023qwen, li2024llama, liu2023visual, liu2024improved, zhu2023minigpt, du2022glm, li2023blip, zhang2023internlm, bavishi2023fuyu, zhang2024jointly, team2024gemini} have achieved remarkable success in multimodal understanding by integrating high-capacity visual encoders with powerful Large Language Models (LLMs)~\cite{touvron2023llama,touvron2023llama2,brown2020language}. However, this superior performance comes at a high computational cost. As modern LVLMs increasingly adopt high-resolution inputs to capture fine-grained spatial details, the volume of visual tokens generated by the encoder grows significantly. This influx of tokens leads to prohibitive inference latency and high memory consumption, posing a substantial challenge for deploying these models in real-time or resource-constrained applications.

Existing efforts~\cite{zheng2023cgc, huang2024ivtp, dhouib2025pact, alvar2025divprune, ye2025atp,chen2024image,zhang2024sparsevlm,zhang2025beyond,zheng2025lightweight} to mitigate this overhead primarily focus on visual token pruning or merging based on heuristic scoring rules. These methods typically rely on internal model statistics, such as attention weights from the vision transformer or feature similarity, to identify and remove redundant tokens. However, such approaches often suffer from two critical limitations. First, they are generally task-agnostic, meaning the pruning decisions are disconnected from the semantic requirements of the downstream language generation task. Second, most existing methods treat pruning as a one-shot operation, failing to model the progressive refinement of visual information as a sequential decision process with step-to-step dependencies.

To overcome these limitations, we propose TPRL, a novel reinforcement learning framework that learns adaptive pruning trajectories through language-guided sequential optimization. Unlike previous works, TPRL formulates token reduction as a Markov Decision Process (MDP) where the pruning policy is directly tied to end-task rewards. To make RL training efficient, we utilize a self-supervised autoencoder to compress visual tokens into a compact state representation. Furthermore, we address the challenge of inefficient exploration by initializing the policy through Learning from Demonstrations (LfD)~\cite{hester2018deep} followed by Proximal Policy Optimization (PPO). This approach ensures that the retained tokens are not only visually prominent but also semantically essential for answering specific textual queries.

Our main contributions are summarized as follows:
\begin{itemize}
\item[$\bullet$]We propose TPRL, a novel reinforcement learning-based framework for visual token pruning in LVLMs, which treats token reduction as a language-guided sequential decision process.
\item[$\bullet$]We introduce an efficient policy learning strategy for token pruning that combines a self-supervised autoencoder for state compression and a Learning from Demonstrations (LfD) phase to improve sample efficiency and stability.
\item[$\bullet$]Extensive experiments demonstrate that TPRL can remove up to 66.7\% of visual tokens and reduce FLOPs by 54.2\% during inference, while incurring only a marginal average accuracy drop of 0.7\% across multiple benchmarks.
\end{itemize}

\section{Related Work}
\label{RW}
\subsection{Large Vision-Language Models}
Large Vision-Language Models (LVLMs), such as LLaVA and its successors~\cite{brown2020language, touvron2023llama, touvron2023llama2, radford2021learning, achiam2023gpt, workshop2022bloom, chiang2023vicuna, zheng2026llava, Zheng_2026}, have achieved state-of-the-art performance by aligning visual features from a pre-trained Vision Transformer (ViT) with the embedding space of a Large Language Model (LLM)~\cite{zheng2025efficient, bai2023qwen, zhu2023minigpt, dai2023instructblip, li2023blip, liu2023visual, zhang2023internlm, wang2024qwen2}. While these models excel in multimodal reasoning and instruction following, their inference efficiency is severely hampered by the massive number of visual tokens. As higher resolutions become standard to capture fine-grained spatial information, the computational burden on the LLM backbone increases significantly. Our work addresses this challenge by introducing a token-level optimization framework that reduces the input density for the LLM without requiring architectural changes to the underlying models.

\subsection{Visual Token Pruning}
Visual token pruning aims to identify and remove redundant or uninformative visual tokens to accelerate model inference. Existing methods typically rely on heuristic scoring mechanisms such as using attention weights from the vision encoder or calculating cosine similarity between visual token embeddings to determine importance. However, these strategies often fail to explicitly model the relationship between the visual token pruning rate and the resulting performance fluctuations as the compression level increases. In contrast, TPRL models visual token pruning as a multi-step decision process with sequential dependencies which allows the policy to better learn the complex priority relationships between different visual tokens. This formulation enables the pruner to progressively refine the visual sequence and adaptively determine the optimal pruning trajectory for maintaining high reasoning accuracy across diverse tasks.

\subsection{Reinforcement Learning for Model Compression}
Reinforcement Learning (RL) has been widely explored for various model compression tasks, including neural architecture search (NAS), channel pruning, and dynamic layer skipping. These approaches treat compression as a decision-making process to find the optimal trade-off between efficiency and accuracy. However, applying RL to visual token pruning in the context of LVLMs is relatively under-explored due to the vast action space and the complexity of multimodal interactions. Some works~\cite{lu2025reinforcementlearningbasedtokenpruning,cao2024madtpmultimodalalignmentguideddynamic} introduce RL to token pruning tasks but few extend it to LVLMs token pruning. TPRL fills this gap by formulating the pruning process as a sequential Markov Decision Process (MDP). By leveraging Learning from Demonstrations (LfD) to provide a warm-start and Proximal Policy Optimization (PPO) for fine-tuning, we effectively navigate the exploration challenges and align the pruning policy with actual task rewards.

\section{Methodology}
\label{sec:method}

\subsection{Overview}
\begin{figure*}[!t]
    \centering
    \includegraphics[width=1\linewidth]{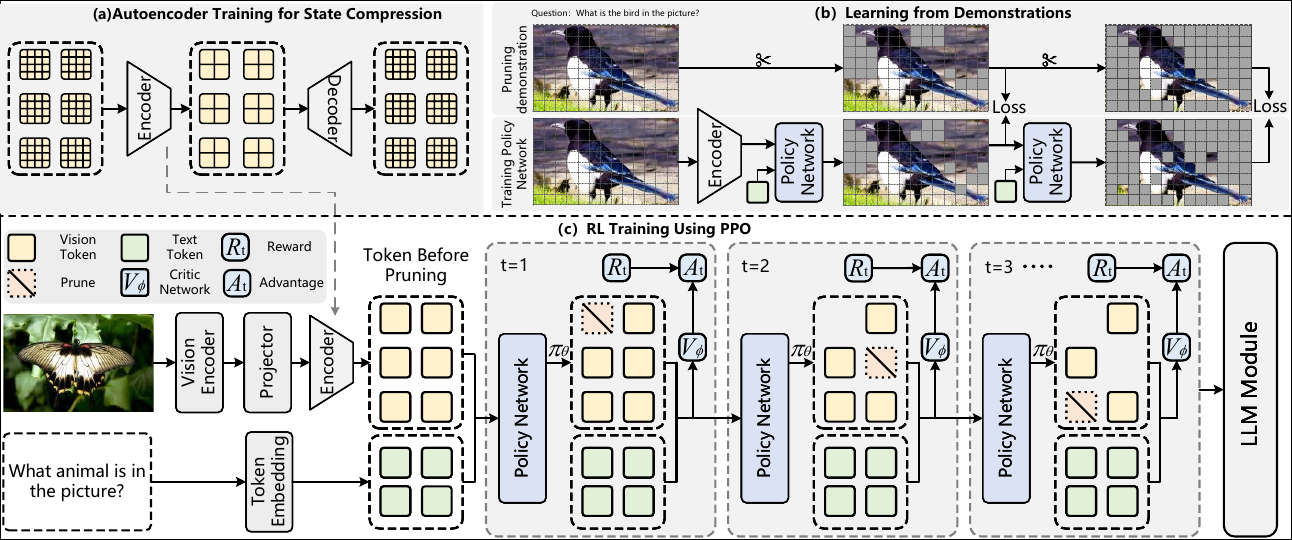}
\caption{Overview of the TPRL framework consisting of three synergistic stages. (a) Autoencoder training for state compression learns a compact latent representation to reduce the feature dimensionality of each visual token. (b) Learning from demonstrations provides a foundational initialization for the pruning policy by imitating high-quality trajectories to ensure stable convergence. (c) RL training using PPO fine-tunes the policy through language-guided sequential optimization to achieve an optimal balance between task performance and computational efficiency.}
    \label{RL-VLM-pruning}
\end{figure*}

In this section, we present TPRL, a reinforcement learning framework for visual token pruning in LVLMs. The overall architecture of our method is shown in Figure~\ref{RL-VLM-pruning}. First, we formulate the adaptive pruning task as a Markov Decision Process (MDP) (Sec.~\ref{sec:MDP}) to enable sequential decision-making. This formulation defines a compact state representation, action space, and reward function. Second, we conduct \emph{Learning from Demonstrations} (Sec.~\ref{sec:lfd}). In this stage, the policy network is pre-trained with supervised learning on pruning trajectories generated by existing heuristic methods. Third, we detail the RL-based adaptive pruning framework (Sec.~\ref{sec:RL}), including the architecture of the policy and value networks. Finally, we present the full training and inference protocol (Sec.~\ref{sec:TAI}), employing Proximal Policy Optimization (PPO)~\cite{schulman2017proximal} for policy optimization.

\subsection{Language-guided MDP Formulation}
\label{sec:MDP}

We formulate the adaptive token pruning task as a Markov Decision Process (MDP) to enable sequential decision-making. The MDP is defined by the tuple $(\mathcal{S}, \mathcal{A}, \mathcal{P}, \mathcal{R}, \gamma)$, where $\mathcal{S}$, $\mathcal{A}$, $\mathcal{P}$, $\mathcal{R}$ represent the state space, action space, state transition dynamics, and reward function, respectively, and $\gamma \in [0,1]$ is the discount factor. 

\noindent\textbf{State Space.}
To enable efficient reinforcement learning, we learn a compact state representation by reducing the feature dimensionality of each visual token independently. We train a lightweight token-wise autoencoder. The encoder $\mathcal{E}$ maps a single visual token $\mathbf{h}_i \in \mathbb{R}^{d_v}$ to a low-dimensional code $\mathbf{z}_i \in \mathbb{R}^{d_l}$ with $d_l \ll d_v$. The decoder $\mathcal{D}$ reconstructs the original token from $\mathbf{z}_i$, and the autoencoder is trained with a per-token mean squared reconstruction objective:
\begin{equation}
\label{eq:recon_loss}
\mathcal{L}_{\text{rec}} = \frac{1}{N} \sum_{i=1}^{N} \|\mathbf{h}_i - \mathcal{D}(\mathbf{z}_i)\|_2^2.
\end{equation}

This compression only reduces feature dimensionality and does not alter the number of visual tokens. After training, we freeze $\mathcal{E}$. At inference and during RL, we apply $\mathcal{E}$ once to obtain the compressed set $\mathbf{Z} = [\mathbf{z}_1,\dots,\mathbf{z}_N] \in \mathbb{R}^{N\times d_l}$. The state at pruning step $t$ is defined as $s_t = (Z_t, q)$, where $Z_t \in \mathbb{R}^{K_t\times d_l}$ contains codes for the $K_t$ remaining tokens and $q \in \mathbb{R}^{d_q}$ denotes the text query embedding. The compressed codes serve only the pruning policy.

We introduce a binary mask vector $\mathbf{m}\in{0,1}^N$ to track the retention of the original visual tokens $\mathbf{H} = [\mathbf{h}_1,\dots,\mathbf{h}_N]^\top \in \mathbb{R}^{N \times d_v}$. The mask is initialized to all ones. After the policy outputs actions at step $t$, any token with original index $idx$ that is marked to be pruned has its corresponding mask entry set to zero: $\mathbf{m}[idx] = 0$. The original indices remain fixed even when earlier tokens are removed. During multi-step RL, the policy always operates on compressed codes, and intermediate masks are applied to the compressed representations for subsequent policy steps, while all mask updates refer to original indices. At inference, the policy produces a one-shot mask by thresholding the retention probabilities, and this mask is applied directly to the original uncompressed tokens. Formally, the retained visual tokens after pruning are given by $\tilde{\mathbf{H}} = { \mathbf{h}_{idx} \mid \mathbf{m}[idx] = 1 }$, which are then forwarded to the downstream LLM. Compressed codes are used only by the pruning policy and are never sent to the LLM.

\noindent\textbf{Action Space.} At each step $t$, the agent outputs a binary retention decision for each token in $Z_t$:
\begin{equation}
a_t = \{d_i \mid i \in \{1, \dots, K_t\}, d_i \in \{0,1\}\},
\end{equation}
where $d_i = 0$ denotes pruning token $i$, and $d_i = 1$ denotes retaining it. The state transition follows:
\begin{equation}
Z_{t+1} = \{\mathbf{z}_{t,i} \mid d_i = 1\},
\end{equation}
where $\mathbf{z}_{t,i}$ is the compressed feature of the $i$-th token at step $t$.

\noindent\textbf{Reward Function.} The reward at step $t$ balances task performance and computational efficiency, computed over a mini-batch of $B$ samples to reduce variance:
\begin{equation}
r_t = R_{\text{task}} + R_{\text{eff}}(t),
\end{equation}
\begin{equation}
R_{\text{task}} = \alpha \cdot \frac{1}{B} \sum_{b=1}^{B} \left[ \mathrm{Score}(H_{t}^{(b)}, q^{(b)}) - \mathrm{Score}(H_{t-1}^{(b)}, q^{(b)}) \right],
\end{equation}
\begin{equation}
R_{\text{eff}}(t) = \beta \cdot \frac{1}{B} \sum_{b=1}^{B} \left(1 - \frac{K_{t}^{(b)}}{K_{t-1}^{(b)}}\right),
\end{equation}
where $\alpha$ and $\beta$ are weighting coefficients, and $\mathrm{Score}$ evaluates the downstream task performance given the retained tokens. Here $H_{t}^{(b)}$ denotes the set of original visual tokens that remain after applying the pruning mask at step $t$ to the initial visual features $H^{(b)}$ (i.e., $H_{t}^{(b)} = \{ \mathbf{h}_i^{(b)} \mid \mathbf{m}_t^{(b)}[i] = 1 \}$).

\subsection{Learning from Demonstrations}
\label{sec:lfd}
The policy network faces the challenge of inefficient exploration and may converge to suboptimal policies when trained from scratch. Inspired by~\cite{chen2021decision}, we pre-train it using supervised learning on high-quality pruning demonstrations. These demonstrations are generated offline using existing heuristic pruning methods that provide reasonable, though sub-optimal, pruning trajectories.

\noindent\textbf{Demonstration Generation.} For each image–text pair in the training set we run a heuristic pruner to produce a short pruning trajectory. Each trajectory $\tau$ is recorded as a sequence of state–action pairs $(s_t,a_t)$, where the state $s_t$ includes the compressed visual representation $Z_t$ and the textual query $q$, and $a_t$ is the binary pruning action recommended by the heuristic at step $t$. Trajectories are generated by applying the reference pruner with progressively increased pruning rates, thereby producing stepwise demonstrations that remove more tokens at each step. Demonstration labels are assigned per token as
\begin{equation}
a^{\text{demo}}_{t,i} =
\begin{cases}
1, & \text{if token } i \text{ is retained at step } t\ ,
\\[4pt]
0, & \text{otherwise}.
\end{cases}
\label{eq:demo_label}
\end{equation}

\noindent\textbf{Supervised Pre-training.} The policy network $\pi_\theta$ (with parameters $\theta$) is trained to imitate the demonstrated actions by minimizing the binary cross-entropy loss over all tokens and steps in the demonstration dataset $\mathcal{D}_{\text{demo}}$:
\begin{equation}
\begin{aligned}
\mathcal{L}_{\text{BCE}}(\theta)
&= -\frac{1}{|\mathcal{D}_{\text{demo}}|}
\sum_{\tau \in \mathcal{D}_{\text{demo}}}
\Bigg[
\sum_{t}
\sum_{i=1}^{K_t}
\Big(
a_{t,i}^{\text{demo}} \log p_{t,i} \\
&\qquad\qquad\qquad
+ (1 - a_{t,i}^{\text{demo}})\log (1 - p_{t,i})
\Big)
\Bigg]
\end{aligned}
\end{equation}
where $a_{t,i}^{\text{demo}}$ is the demonstration action for token $i$ at step $t$, and $p_{t,i} = \pi_\theta(s_t)_i$ is the retention probability predicted by the policy network. This supervised pre-training phase equips the policy with a foundational understanding of token relevance, significantly improving sample efficiency and stability during the subsequent RL fine-tuning.

\subsection{RL-based Adaptive Token Pruning}
\label{sec:RL}

Our RL framework leverages the pre-trained policy network as a starting point and fine-tunes it using PPO to optimize the pruning strategy. The policy and value networks share an attention-based feature extractor~\cite{schulman2017proximal,hirsch2022multi} that operates on the compressed state $s_t$. This reduces the number of parameters, improves training efficiency, and reduces memory usage.

\noindent\textbf{Shared Attention Module.} The module processes the concatenated sequence of compressed visual tokens and the projected query:
\begin{equation}
X = [Z_t; \tilde{q}] \in \mathbb{R}^{(K_t+1) \times d},
\end{equation}
where $\tilde{q}$ is the linearly projected query embedding. Multi-head self-attention with residual connections models token relationships:
\begin{equation}
\text{Attention}(X) = \text{softmax}\left(\frac{XW_Q(XW_K)^\top}{\sqrt{d_k}}\right)XW_V,
\end{equation}
\begin{equation}
F = \text{LayerNorm}(X + \text{Concat}(\text{head}_1,\dots,\text{head}_h)W_O),
\end{equation}
producing enriched features $F = [f_1, \dots, f_{K_t}, f_{q}]$.

\noindent\textbf{Policy Network.} For each visual token $i$, the policy network computes a retention probability:
\begin{equation}
g_i = \text{LayerNorm}(f_i + W_2 \cdot \text{GELU}(W_1 f_i + b_1) + b_2),
\end{equation}
\begin{equation}
p_{t,i} = \sigma(w^\top g_i + b).
\end{equation}
The action for each visual token is sampled from a Bernoulli distribution whose success probability is discounted by a step-wise factor during RL:
\begin{align}
a_{t,i} &\sim \mathrm{Bernoulli}\big(\lambda_{\mathrm{disc}}^{t} p_{t,i}\big), \label{eq:bernoulli_sample_lambda_disc}\\
P(a_{t,i}=1) &= \lambda_{\mathrm{disc}}^{t} p_{t,i}, \quad
P(a_{t,i}=0) = 1 - \lambda_{\mathrm{disc}}^{t} p_{t,i}, \label{eq:bernoulli_probs_lambda_disc}
\end{align}
where $\lambda_{\mathrm{disc}}\in(0,1)$ is a step-wise discount factor that decays the retention probability with step index $t$, encouraging more aggressive pruning in later steps and accelerating training. We sample actions during RL for exploration and gradient estimation. 

At inference we use deterministic one-shot pruning:
\begin{equation}
a_{t,i} = \mathbf{1}\{p_{t,i}>\tau\}. \label{eq:inference_action_lambda_disc}
\end{equation}

\noindent\textbf{Value Network.} The value network estimates the state value by aggregating token features:
\begin{equation}
\bar{f} = \frac{1}{K_t} \sum_{i=1}^{K_t} f_i,
\end{equation}
\begin{equation}
V_\phi(s_t) = u^\top \text{LayerNorm}(\bar{f} + U_2 \cdot \text{GELU}(U_1 \bar{f} + c_1) + c_2) + b_v.
\end{equation}

\subsection{Training and Inference}
\label{sec:TAI}

\noindent\textbf{Training Strategy.} Our training proceeds in three phases: (1) Autoencoder training for state compression via $\mathcal{L}_{\text{rec}}$; (2) Supervised policy pre-training via $\mathcal{L}_{\text{BCE}}$ using demonstrations; (3) RL fine-tuning using PPO. During RL, we collect trajectories over $T_{\max}$ pruning steps per sample. The advantage $\hat{A}_t$ is estimated using Generalized Advantage Estimation (GAE)~\cite{schulman2015high}:
\begin{equation}
\delta_t = r_t + \gamma V_{\phi_{\text{old}}}(s_{t+1}) - V_{\phi_{\text{old}}}(s_t)
\end{equation}
\begin{equation}
\quad \hat{A}_t = \sum_{l=0}^{T-t-1} (\gamma\lambda)^l \delta_{t+l}.
\end{equation}
The PPO objective combines clipped policy loss, value loss, and entropy regularization:
\begin{equation}
L^{\text{CLIP}}(\theta) = \mathbb{E}_t\left[ \min\left( r_t(\theta) \hat{A}_t,\ \text{clip}\big(r_t(\theta), 1-\epsilon, 1+\epsilon\big)\hat{A}_t \right) \right],
\end{equation}
\begin{equation}
L^{\text{VF}}(\phi) = \mathbb{E}_t\left[\big(V_\phi(s_t) - V_t^{\text{target}}\big)^2\right],
\end{equation}
\begin{equation}
L^{\text{S}}(\theta) = \mathbb{E}_t\left[ \mathcal{H}(\pi_\theta(\cdot|s_t)) \right].
\end{equation}
The total optimization objective is then defined as:
\begin{equation}
L(\theta,\phi) = -L^{\text{CLIP}}(\theta) + c_1 L^{\text{VF}}(\phi) - c_2 L^{\text{S}}(\theta),
\end{equation}
where $r_t(\theta) = \pi_\theta(a_t|s_t) / \pi_{\theta_{\text{old}}}(a_t|s_t)$ is the probability ratio between the new and old policies, $\mathcal{H}$ denotes the entropy of the policy distribution, and $c_1$ and $c_2$ are weighting coefficients that balance the value regression loss and entropy regularization, respectively.

\noindent\textbf{Inference Protocol.} During inference, we perform one-shot pruning: the policy network processes all compressed visual tokens once, outputs retention probabilities $\{p_i\}$, and prunes tokens where $p_i \leq \tau$. The threshold $\tau$ is adjustable to trade off efficiency and accuracy. The retained tokens are passed to the downstream LVLM for task execution.

\section{Experiments}
\label{exp}

\subsection{Experimental Setup}

\subsubsection{Implementation Details}
\begin{wrapfigure}[12]{r}{0.42\columnwidth}
\vspace{-8mm}
\centering
\includegraphics[width=\linewidth]{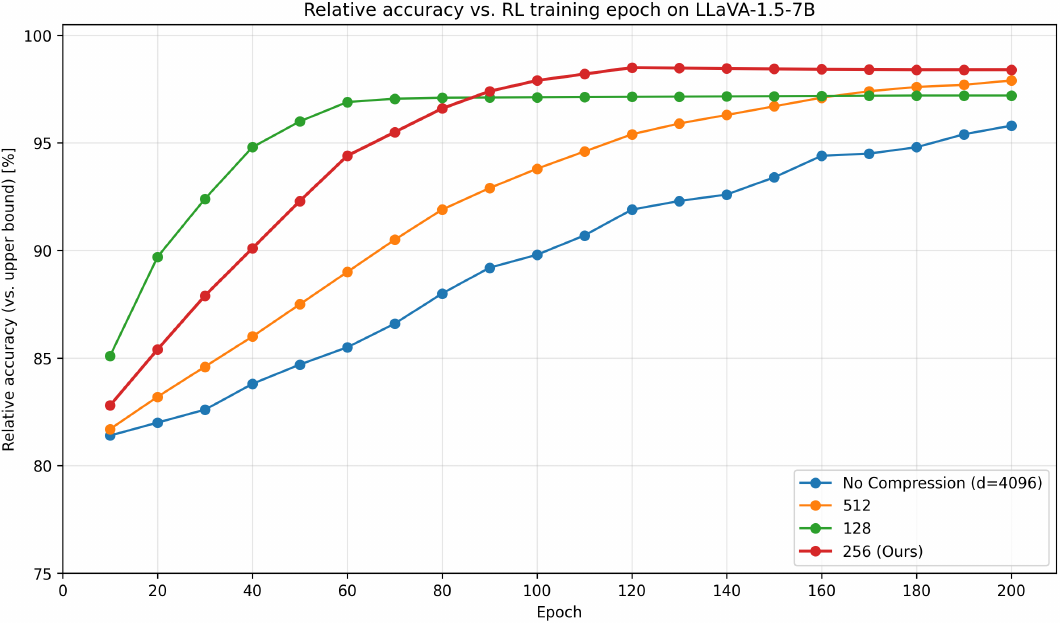}
\caption{Impact of state compression dimension $d_l$ on relative accuracy across RL training iterations.}
\label{fig:compression}
\end{wrapfigure}
We implement our framework on top of LLaVA-1.5~\cite{liu2024improved} and Qwen-VL-2.5~\cite{bai2023qwen}. For LLaVA-1.5, we adopt both the 7B and 13B language models with CLIP-ViT-L/14 as the vision encoder while for Qwen-VL-2.5 we use the official 3B model with NaViT~\cite{dehghani2023patch}. The token-wise autoencoder for state compression consists of a two-layer MLP encoder and a symmetric decoder where the encoder reduces the feature dimensionality of each visual token to 1/16 of its original size by mapping 1024-dimensional embeddings through a hidden layer of 256 units with LayerNorm and GELU activation to a 64-dimensional latent space. The decoder mirrors this structure to reconstruct the original features, after which the encoder is frozen for all subsequent stages. 
To generate demonstrations for pre-training, we utilize Vispruner~\cite{zhang2025beyond} to construct 50K pruning trajectories from the training data. For each image–text pair, we generate a short trajectory by applying two pruning rates sequentially from no mask to a 25\% pruning mask and finally to a 50\% pruning mask. These short demonstration trajectories are intentionally used to accelerate policy convergence by providing a reasonable initialization for visual token relevance. The subsequent RL stage then refines this initialization and learns longer pruning sequences beyond the demonstrated steps to optimize for final task rewards.

The policy network is pre-trained to imitate these demonstrations. After supervised pre-training, we fine-tune the policy with PPO. PPO training uses discount factor $\gamma=0.99$, GAE parameter $\lambda=0.95$, and reward weights $\alpha=1.0$ (task reward) and $\beta=0.1$ (efficiency reward). The PPO objective uses value loss coefficient $c_1=0.5$ and entropy regularization coefficient $c_2=0.01$. During RL fine-tuning, we collect trajectories with up to $T_{\max}=3$ pruning steps per sample, allowing the policy to explore and learn longer sequential pruning strategies than those present in the demonstrations. We set $\lambda_{\mathrm{disc}}=0.5$. At inference time, we perform one-shot pruning: The policy outputs per-token retention probabilities. At inference we prune tokens with $p_i \le \tau$. In our experiments (averaged across benchmarks), $\tau=0.55$ yields about 192 retained visual tokens, $\tau=0.67$ yields about 144 tokens, and $\tau=0.74$ yields about 128 tokens.
\begin{wrapfigure}[13]{r}{0.5\columnwidth}
\vspace{-6mm}
\centering
\includegraphics[width=\linewidth]{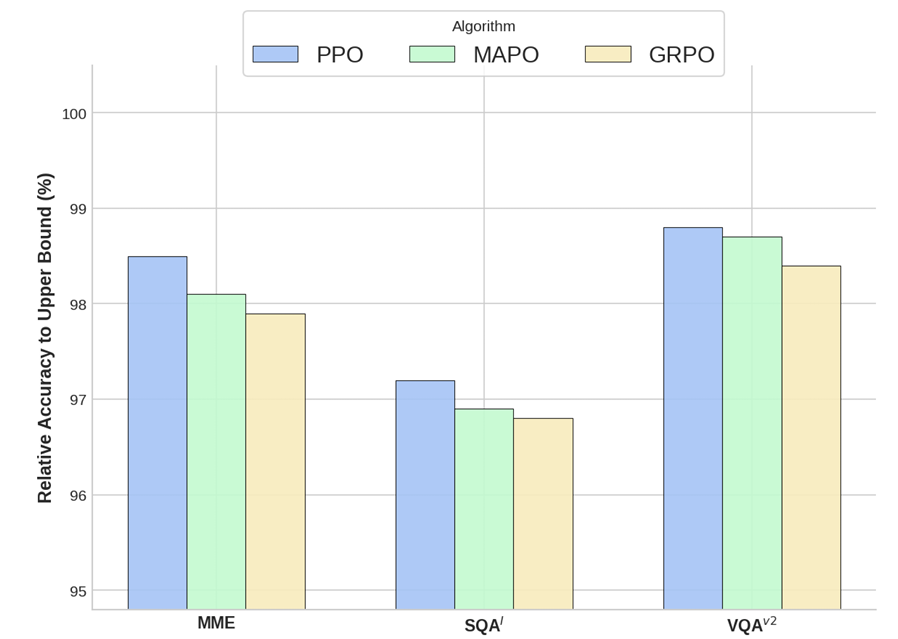}
\caption{Comparison of RL optimization methods.}
\label{fig:rl_comparison}
\end{wrapfigure}

Further details on datasets and training configurations (e.g., learning rate, optimizer, GPU setup) are provided in the \textcolor{red}{Supplementary Materials}. Ablation studies on hyperparameters (e.g., the scaling coefficients for each loss term) will also be presented in the supplementary materials.

\vspace{-1em}
\subsubsection{Datasets and Metrics}

We evaluate our method on three upper bound models: LLaVA-1.5-7B, LLaVA-1.5-13B, and Qwen2.5-VL-3B. For LLaVA-1.5-7B, we conduct experiments on all seven benchmarks: GQA~\cite{hudson2019gqa}, MMBench~\cite{liu2024mmbench}, MME~\cite{fu2023mme}, POPE~\cite{li2023evaluating}, SEED-Bench~\cite{li2023seed}, ScienceQA-Image~\cite{lu2022learn}, and VQA-v2~\cite{goyal2017making}. For LLaVA-1.5-13B, we use a subset comprising GQA, MMBench, MME, POPE, SEED-Bench, and ScienceQA-Image. For Qwen2.5-VL-3B, we evaluate on GQA, MME, ScienceQA-Image, and VQA-v2. These benchmarks cover a wide range of capabilities including visual question answering, reasoning, perception, and hallucination evaluation. For training the token-wise autoencoder, generating pruning demonstrations, and supervised pre-training of the policy, we use the training splits of all seven datasets to maximize coverage of visual concepts and question styles. We report absolute task performance on each benchmark as well as relative percentage with respect to the corresponding upper bound model (i.e., the full model using all visual tokens without pruning) to facilitate a fair efficiency–accuracy trade-off analysis.
\subsection{Experimental Results}

\noindent\textbf{Main Results.} We evaluate TPRL across multiple benchmarks by adjusting the threshold $\tau$ to align average visual token retention to 192, 144, and 128, which correspond to average pruning rates of 66.7\%, 75\%, and 77.8\%, respectively. The comprehensive experimental results for LLaVA-1.5-7B, LLaVA-1.5-13B, and Qwen-2.5-VL are summarized in Table~\ref{tab:main_results_llava}, Table~\ref{tab:main_results_LLaVA13B}, and Table~\ref{tab:main_results_qwen3b}. For LLaVA-1.5-7B, as shown in Table~\ref{tab:main_results_llava}, TPRL achieves nearly lossless performance with a marginal 0.7\% average accuracy drop at 192 visual tokens while maintaining competitive drops of 1.6\% and 2.3\% at higher compression levels. Notably, TPRL exceeds the Upper Bound on MMBench because the RL agent effectively filters redundant visual noise and reduces potential interference for the LLM. Our method also demonstrates superior scalability on LLaVA-1.5-13B, where the data in Table~\ref{tab:main_results_LLaVA13B} show that average performance drops by only 0.8\% and 2.9\% at 192 and 128 visual tokens. Similar robustness is observed on Qwen-2.5-VL in Table~\ref{tab:main_results_qwen3b}, which incurs manageable drops of 1.9\% and 2.3\% at the same retention levels. These consistent findings across diverse model architectures and scales confirm the task-awareness and generalizability of the TPRL framework.

\begin{table*}[!t]
  \centering
  \fontsize{6pt}{6pt}\selectfont  
  \caption{Performance comparison on multiple benchmarks with the LLaVA-1.5-7B backbone. We report absolute scores and the compression retention rates to Upper Bound model. (Token) denotes the average visual token count per LLM layer. Methods marked with (*) means the retained token count is adaptively determined and we calculate the average token count during inference across all benchmarks.}
  \begin{tabularx}{1\textwidth}{l|>{\centering\arraybackslash}p{1cm}|*{7}{>{\centering\arraybackslash}X}|>{\centering\arraybackslash}p{0.8cm}}
    \toprule[1pt]
    Method  & Token & \textbf{GQA} & \textbf{MMB} & \textbf{MME} & \textbf{POPE} & \textbf{SEED} & \textbf{SQA$^{I}$} & \textbf{VQA$^{v2}$} & \textbf{Avg.} \\
    \midrule[1pt]
    
    \multicolumn{10}{c}{\textbf{Upper Bound Model}} \\
    \midrule
    
    \multirow{2}{*}{LLaVA-1.5-7B~\cite{liu2024improved}} & \multirow{2}{*}{576} & 62.0 (100\%) & 64.6 (100\%) & 1553.4 (100\%) & 85.9 (100\%) & 60.3 (100\%) & 70.4 (100\%) & 78.9 (100\%) & - (100\%) \\
     
    \midrule

    \multicolumn{10}{c}{\textbf{Methods without RL}} \\
    \midrule
    
    \multirow{4}{*}{{ToMe~\cite{bolya2022token}}} & \multirow{2}{*}{192}   & {54.3 (87.7\%)} & {60.5 (93.7\%)} & {1485 (95.6\%)} & {72.4 (84.3\%)} & {53.1 (88.1\%)} & {65.2 (92.6\%)} & {67.9 (86.1\%)} & - (90.5\%) \\
      \cmidrule{2-10}
     & \multirow{2}{*}{128}   & {52.4 (84.5\%)} & {53.3 (82.5\%)} & {1343 (86.5\%)} & {62.8 (73.1\%)} & {50.9 (84.4\%)} & {59.6 (84.7\%)} & {63.0 (79.8\%)} & - (85.4\%) \\
    
    \midrule

    \multirow{4}{*}{{PruMerge+~\cite{shang2024LLaVA-PruMerge}}} & \multirow{2}{*}{144}   &61.3 (98.7\%)& 64.9 (100.5\%) & 1462.4 (94.1\%) & 84.0 (97.8\%) & {-} & 68.3 (97.0\%) & \textbf{76.8} (97.3\%) & - (97.6\%) \\
      \cmidrule{2-10}
     & \multirow{2}{*}{128}   & {57.8 (93.2\%)} & {61.3 (94.9\%)} & {1420.5 (91.4\%)} & {81.5 (94.9\%)} & {-} & {67.6 (96.0\%)} & {74.7 (94.7\%)} & - (96.9\%) \\
    \midrule

    \multirow{4}{*}{{FastV~\cite{chen2024image}}} & \multirow{2}{*}{192}   & 52.7 (85.0\%) & 61.2 (94.7\%) & 1312.5 (84.7\%) & 64.7 (75.3\%) & 50.9 (84.4\%) & 65.4 (92.9\%) & 67.2 (85.2\%) & - (86.0\%) \\
      \cmidrule{2-10}
     & \multirow{2}{*}{128}   & 49.6 (80.0\%) & 56.1 (86.7\%) & 1195.4 (77.0\%) & 59.2 (67.0\%) & 48.5 (90.4\%) & 59.7 (84.8\%) & 61.8 (78.3\%) & - (79.9\%) \\
    \midrule
    
    \multirow{4}{*}{{SparseVLM~\cite{zhang2024sparsevlm}}} & \multirow{2}{*}{192}   & {57.6} (92.3\%) & 62.5 (96.7\%) & {1473.0} (94.8\%) & 83.6 (97.3\%) & 53.0 (97.9\%) & 67.2 (95.5\%) & 75.6 (95.8\%) & - (94.2\%) \\
      \cmidrule{2-10}
     & \multirow{2}{*}{128}   & 56.0 (90.3\%) & 60.0 (92.9\%) & 1359.7 (87.5\%) & 80.5 (93.7\%) & 50.4 (83.6\%) & 65.1 (92.5\%) & 73.2 (92.8\%) & - (91.0\%) \\
     \midrule
    \multirow{4}{*}{{Vispruner~\cite{zhang2025beyond}}} & \multirow{2}{*}{192}   & {58.1} (92.2\%) & 62.9 (97.3\%) & {1475.0} (95.0\%) & 84.9 (98.8\%) & -  & 69.2 (98.3\%) & 76.5 (97.0\%) & - (96.4\%) \\
     \cmidrule{2-10}
     & \multirow{2}{*}{128}   & 57.9 (93.4\%) & 62.7 (97.0\%) & 1460.8 (94.0\%) & 84.5 (98.4\%) & -  & 68.5 (97.3\%) & 75.8 (96.1\%) & - (96.0\%) \\
    
    \midrule
    
    \multicolumn{10}{c}{\textbf{Methods with RL}} \\
    \midrule
    
    \multirow{6}{*}{\textbf{Ours}} & \multirow{2}{*}{192*}  & 61.1 (98.5\%) & 66.1 (102.3\%) &1554.1 (99.4\%) &85.6 (99.7\%) &58.9 (97.6\%) &69.0 (98.0\%) &78.6 (99.6\%) & - (99.3\%) \\
    \cmidrule{2-10}
    & \multirow{2}{*}{144*} &60.7 (97.9\%) &65.8 (101.8\%) &1539.4 (99.1\%) &84.0 (97.8\%) &58.5 (97.0\%) &68.1 (96.8\%) &77.5 (98.2\%) & - (98.4\%) \\
    \cmidrule{2-10}
    & \multirow{2}{*}{128*} &60.2 (97.1\%) & 65.4 (101.2\%) &1533.2 (98.7\%) &83.2 (96.8\%) &58.2 (96.6\%) &67.7 (96.2\%) &76.8 (97.3\%) & - (97.7\%) \\
    
    \bottomrule[1pt]
  \end{tabularx}
\vspace{-3em}
\label{tab:main_results_llava}
\end{table*}

\begin{table*}[t]
  \centering
  \fontsize{6pt}{8pt}\selectfont
  \caption{Performance comparison on multiple benchmarks with LLaVA-1.5-VL-13B backbone.}
  \begin{tabularx}{1\textwidth}{l|>{\centering\arraybackslash}X|*{5}{>{\centering\arraybackslash}X}|>{\centering\arraybackslash}X}
    \toprule[1pt]
    Method  & Token & \textbf{GQA} & \textbf{MMB} & \textbf{MME} & \textbf{POPE} & \textbf{SQA$^{I}$} &  \textbf{Avg.} \\
   \midrule[1pt]
    \multirow{2}{*}{LLaVA-1.5-VL-13B} & \multirow{2}{*}{576} & {63.3} & {68.64} & {1522}  & {85.99} & {72.88}  & {-} \\
     && {100\%} & {100\%}  & {100\%} & {100\%} & {100\%}  & {100\%}\\
    \midrule
    \multirow{4}{*}{FastV~\cite{chen2024image} } & \multirow{2}{*}{192} 
     & {54.2} & {64.8}  & {1290} & {64.1}  & {69.9} & {-} \\
     & & {85.7\%} & {94.5\%} & {84.7\%}  & {75.2\%} & {92.1\%} & {86.4\%}  \\
     \cline{2-8}
     & \multirow{2}{*}{128} 
     & {48.5} & {61.2} & {1150}  & {58.2} & {68.0} & {-} \\
     & & {76.6\%} & {89.2\%}  & {75.6\%} & {67.7\%} & {93.3\%} & {80.5\%} \\
    \midrule
    \multirow{4}{*}{\textbf{Ours} } & \multirow{2}{*}{192*} 
     & {62.8} & {68.95} & {1500}  & {85.5} & {71.8} & {-} \\
     & & {99.2\%} & {100.5\%}  & {98.6\%} & {99.4\%} & {98.5\%} & {99.2\%} \\
     \cline{2-8}
     & \multirow{2}{*}{128*} 
     & {61.2} & {67.5} & {1465}  & {83.5} & {70.8} & {-} \\
     & & {96.7\%} & {98.3\%}  & {96.3\%} & {97.1\%} & {97.1\%} & {97.1\%} \\
   \bottomrule[1pt]
  \end{tabularx}
\label{tab:main_results_LLaVA13B}
\end{table*}

\begin{table*}[t]
  \centering
  \fontsize{6pt}{8pt}\selectfont
  \caption{Performance comparison on multiple benchmarks with Qwen-VL-2.5 backbone.}
  \begin{tabularx}{1\textwidth}{l|>{\centering\arraybackslash}X|*{4}{>{\centering\arraybackslash}X}|>{\centering\arraybackslash}X}
    \toprule[1pt]
    Method  & Token & \textbf{GQA} &\textbf{MME}  & \textbf{SQA$^{I}$} & \textbf{VQA$^{v2}$} & \textbf{Avg.} \\
   \midrule[1pt]
    \multirow{2}{*}{Qwen-2.5-VL-3B~\cite{bai2023qwen}} & \multirow{2}{*}{256}  
     & {61.5} & {1450} & {70.2} & {77.4} & {-} \\
     && {100\%} & {100\%} & {100\%} & {100\%} & {100\%}\\
    \midrule
    \multirow{4}{*}{FastV~\cite{chen2024image} } & \multirow{2}{*}{144} 
     & {54.7} & {1305} & {64.6} & {73.5} & {-} \\
     & & {89.0\%} & {90.0\%} & {92.0\%} & {95.0\%} & {91.5\%} \\
     \cline{2-7}
     & \multirow{2}{*}{128} 
     & {49.5} & {1232} & {61.1} & {67.7} & {-} \\
     & & {80.5\%} & {85.0\%} & {87.0\%} & {87.5\%} & {85.0\%}\\
    \midrule
    \multirow{4}{*}{\textbf{Ours} } & \multirow{2}{*}{144*} 
     & {60.4} & {1399} & {68.2} & {77.9} & {-} \\
     & & {98.2\%} & {96.5\%} & {97.1\%} & {100.6\%} & {98.1\%} \\
     \cline{2-7}
     & \multirow{2}{*}{128*} 
     & {60.1} & {1393} & {67.7} & {77.7} & {-} \\
     & & {97.8\%}  & {96.1\%} & {96.5\%} & {100.4\%} & {97.7\%}\\
   \bottomrule[1pt]
  \end{tabularx}
\label{tab:main_results_qwen3b}
\end{table*}

\begin{table*}[t]
  \centering
   \fontsize{8pt}{8pt}\selectfont 
   \caption{Efficiency analysis including cache storage memory, latency and the FLOPs. \(\Delta\) denotes the reduction ratio.}
  \begin{tabular}{l|c|c|c@{\hspace{0.5em}}c|c@{\hspace{0.5em}}c|c@{\hspace{0.5em}}c}
    \toprule[1pt]
    \multirow{2}{*}{Method} & Avg. & \multirow{2}{*}{Accuracy} &  Storage & \multirow{2}{*}{\(\Delta\)} & Latency  & \multirow{2}{*}{\(\Delta\)} & FLOPs & \multirow{2}{*}{\(\Delta\)} \\
     & Token & & Memory (MB) &  &  (ms) $\downarrow$ &  & (T) $\downarrow$ &  \\
    \midrule[1pt]
    Upper Bound                & 576 & 100\%  & 302.4 & -      & 57.8 & -                      & 9.6 & -    \\
    FastV~\cite{chen2024image} & 192 & 86.0\%   & 100.8  & 66.7\% & 23.1 & 60.0\% &  2.0& 79.2\%    \\
    SparseVLM~\cite{zhang2024sparsevlm} & 192 & 94.2\%   & 100.8 & 66.7\% & 26.0  & 55.0\% & 3.9 & 59.4\%    \\
    \midrule
    
    \multirow{3}{*}{\textbf{Ours}} & 192* & 99.3\%   & 100.9 & 66.6\% & 26.7 & 53.8\% & 4.4 & 54.2\%     \\
     & 144* & 98.4\%   & 75.7 & 75.0\% & 22.6 & 60.9\% & 3.7 & 61.4\%     \\
     & 128* & 97.7\%   & 67.3 & 77.7\% & 21.2 & 63.3\% & 3.5 & 63.5\%     \\
    \bottomrule[1pt]
  \end{tabular}
\vspace{-2em}
\label{tab:efficiency_analysis}
\end{table*}

\noindent\textbf{Efficiency Analysis.} 
We provide a comprehensive evaluation of TPRL by analyzing its cache storage memory, latency, and FLOPs based on the total inference complexity $\mathcal{F}_{\text{total}} = \mathcal{F}_{\text{vision}} + \mathcal{F}_{\text{pruner}} + \mathcal{F}_{\text{prefill}} + \mathcal{F}_{\text{decode}}$ where $\mathcal{F}_{\text{vision}}$ and $\mathcal{F}_{\text{pruner}}$ represent the costs of the vision encoder and our pruning module. The LLM backbone workload comprises the prefill phase $\mathcal{F}_{\text{prefill}} \approx 2P \cdot (N_{v'} + N_t)$ for parallel input processing and the decoding phase $\mathcal{F}_{\text{decode}}$ for autoregressive generation. During the decoding stage, which accounts for a substantial portion of the inference time, reducing $N_{v'}$ directly minimizes the size of the KV cache and decreases the computational cost of the attention mechanism for every newly generated token. As summarized in Table~\ref{tab:efficiency_analysis}, the reduction in visual tokens yields substantial efficiency gains across all execution metrics. While heuristic methods like FastV and SparseVLM achieve zero $\mathcal{F}_{\text{pruner}}$ through simple attention-based dropping, TPRL establishes a superior accuracy-efficiency Pareto frontier by ensuring that only the most semantically relevant visual tokens are retained for the final reasoning task.

\subsection{Ablation Study}

\noindent\textbf{Impact of LLM Fine-tuning.} We evaluate the necessity of LLM fine-tuning by comparing TPRL against a frozen baseline using LLaVA-1.5-7B with 192 tokens. Table~\ref{tab:llm_ft} shows that fine-tuning significantly improves performance by adapting internal representations to sparse visual inputs. This optimization enables the LLM to better interpret critical tokens and mitigate information loss from pruning. Consequently, fine-tuning ensures robust multimodal reasoning despite substantial token reduction.

\noindent\textbf{Impact of State Compression Dimension.} We investigate the impact of the compressed feature dimension $d_l$ on the dynamic performance of TPRL using LLaVA-1.5-7B with the visual token count fixed at 192. As illustrated in Figure~\ref{fig:compression}, reducing the token features from the original 4096 dimensions down to 256 allows the agent to achieve convergence speeds and final accuracies nearly identical to the uncompressed baseline. This dynamic behavior confirms that the pruning policy primarily relies on high-level semantic relevance which can be effectively preserved in a compact latent space. Conversely, a further reduction to 128 dimensions introduces a significant information bottleneck that degrades performance throughout the entire training process. The optimal $16\times$ compression ratio used in TPRL thus provides sufficiently informative state representations to identify semantically critical tokens while ensuring the pruning process remains virtually computationally free across diverse benchmarks.


\begin{wraptable}{r}{0.50\columnwidth}

\vspace{-5mm}

\centering

\caption{Ablation study on LLM fine-tuning.}
\label{tab:llm_ft}
\resizebox{\linewidth}{!}{
\begin{tabular}{lcccc}
\toprule
\textbf{LLM Status}  & \textbf{VQA$^{v2}$} & \textbf{MMB}  & \textbf{MME} & \textbf{POPE}  \\
\midrule
Frozen  & 98.8\%  & 99.5\%  & 98.5\% & 98.6\%   \\
\textbf{Fine-tuned} & 99.6\% & 102.3\%  & 99.4\% &  99.7\% \\
\bottomrule
\end{tabular}
}

\vspace{4mm}

\caption{Ablation study on policy initialization and refinement. Results are reported as the relative percentage with respect to the upper bound.}
\label{tab:LFD}
\resizebox{\linewidth}{!}{
\begin{tabular}{lcccc}
\toprule
\textbf{Method}  & \textbf{VQA$^{v2}$} & \textbf{MMB}  & \textbf{MME} & \textbf{POPE}  \\
\midrule
Vispruner (Static) & 97.0\%  & 97.1\% & 95.0\%  & 98.8\%   \\
RL from scratch & 96.3\% & 98.5\% & 96.7\% & 95.9\%   \\
\textbf{TPRL (Full)} & 99.6\% & 102.3\% & 99.4\%  & 99.7\%  \\
\bottomrule
\end{tabular}
}

\vspace{-6mm}

\end{wraptable}


\noindent\textbf{Effect of Policy Initialization and Refinement.} We evaluate the impact of our policy initialization and subsequent refinement on LLaVA-1.5-7B with the visual token count fixed at 192. As shown in Table~\ref{tab:LFD}, we compare the full TPRL framework against the static Vispruner heuristic and a policy network trained from scratch. Although Vispruner provides a stable starting point, its static nature prevents it from adapting to the diverse semantic requirements of different downstream tasks. Conversely, training the policy from scratch leads to poor convergence and suboptimal results because the agent fails to navigate the vast token-wise action space without a prior semantic warm start. TPRL addresses these limitations by using Vispruner trajectories to provide a robust policy initialization. This demonstration-guided foundation enables the subsequent reinforcement learning phase to effectively explore and discover superior adaptive pruning strategies which significantly outperform both the heuristic baseline and the randomly initialized policy across all benchmarks.



\noindent\textbf{Maximum Pruning Steps Analysis.} 

We evaluate the impact of the maximum pruning steps $T_{\max}$ during RL training by varying it from 1 to 5 on the VQA-v2 benchmark on LLaVA-1.5-7B with 192 visual tokens. The training epoch is set as 200.
Table~\ref{tab:tmax} shows that accuracy improves as $T_{\max}$ increases from 1 to 3 due to the richer sequential decision-making context provided during policy learning for visual token selection. Accuracy plateaus beyond $T_{\max}=3$ which justifies our choice of $T_{\max}=3$ to balance policy performance and training complexity.

\begin{wraptable}{r}{0.50\columnwidth}

\vspace{-10mm}

\centering

\caption{Impact of $T_{\max}$ on VQA-v2.}
\label{tab:tmax}
\resizebox{\linewidth}{!}{
\begin{tabular}{lccccc}
\toprule
$T_{\max}$ & 1 & 2 & 3 & 4 & 5 \\ \midrule
Rel. Acc (\%) & 95.2 & 98.8 & 99.6 & 96.2 & 93.7 \\ \bottomrule
\end{tabular}
}

\vspace{4mm}

\caption{Ablation study on PolicyNet architecture.}
\label{tab:policy_arch}
\resizebox{\linewidth}{!}{
\begin{tabular}{lcccc}
\toprule
\textbf{Architecture}  & \textbf{VQA$^{v2}$} & \textbf{MMB}  & \textbf{MME} & \textbf{POPE}  \\
\midrule
MLP-only  & 87.4\%  & 89.6\% & 87.9\% & 89.3\%  \\
\textbf{Attn. + MLP (Ours)} & 99.6\% & 102.3\% & 99.4\% & 99.7\%   \\
\bottomrule
\end{tabular}
}

\vspace{-4mm}

\end{wraptable}




\noindent\textbf{Impact of PolicyNet Architecture.} We evaluate the necessity of attention-based contextual modeling by comparing our design against a baseline global MLP architecture on LLaVA-1.5-7B with 192 visual tokens. As shown in Table~\ref{tab:policy_arch}, the global MLP structure yields inferior pruning precision because it struggles to effectively capture the complex relational dependencies and redundancies between visual tokens. In contrast, our self-attention module explicitly enables each token to interact with the entire visual field to generate a highly adaptive pruning mask. This holistic observation is essential for the policy to accurately distinguish between task-critical information and disposable visual noise across diverse semantic contexts.

Consequently, the attention-based architecture significantly improves performance across all benchmarks while maintaining negligible computational overhead through our compact state representation.

\noindent\textbf{Reinforcement Learning Method Comparison.} 
We assess the robustness of TPRL to various RL optimizers by comparing PPO, MAPO~\cite{liang2018memory}, and GRPO~\cite{vojnovic2025alignment} on LLaVA-1.5-7B with 192 visual tokens. As shown in Figure~\ref{fig:rl_comparison}, PPO marginally outperforms MAPO and GRPO while all three methods maintain nearly comparable accuracy across benchmarks. This minimal performance gap suggests that our framework is highly insensitive to the specific optimization algorithm because the LfD initialization already establishes a strong semantic foundation for the policy. Consequently, PPO is adopted as the default optimizer due to its slightly superior stability even though other standard policy gradient variants achieve similar results. This robustness confirms that the efficacy of TPRL primarily stems from our task-aware MDP formulation and demonstration-guided warm start rather than a specific optimization technique.

\begin{figure*}[!t]
\begin{minipage}[b]{1.0\linewidth}
  \centering
  \centerline{\includegraphics[width=11.5cm]{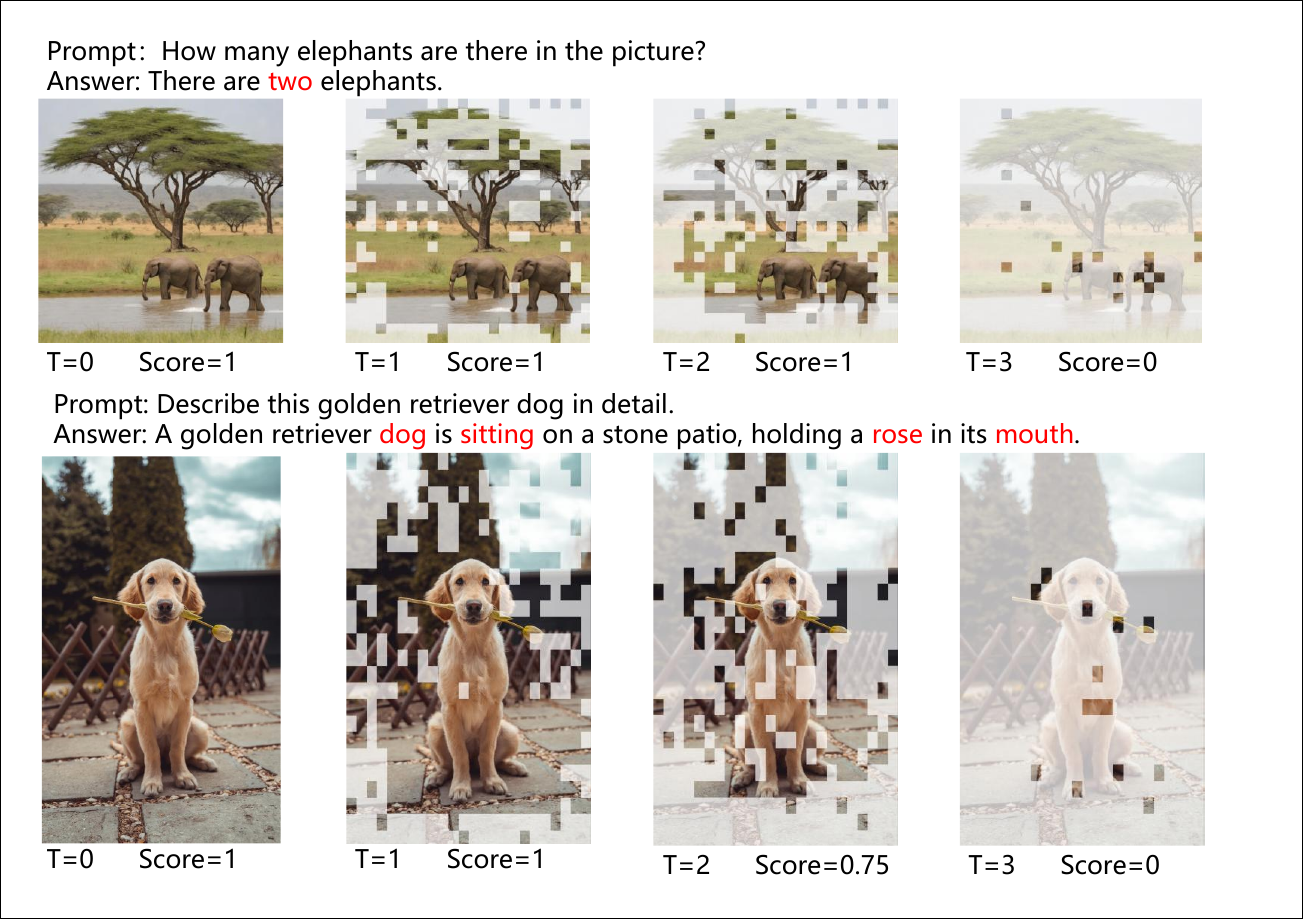}}
\end{minipage}
\caption{Qualitative visualization of sequential pruning trajectories via mask superimposition. The agent progressively filters backgrounds and redundant details to isolate task-critical information through our multi-step MDP formulation.}
\vspace{-2em}
\label{fig:RLPruningVisualization}
\end{figure*}

\noindent\textbf{Visualization Results.} 
Figure~\ref{fig:RLPruningVisualization} qualitatively illustrates the sequential pruning trajectories via mask superimposition. In this visualization, $T$ denotes the decision step from the initial state $T=0$, and the Score reflects the LLM performance after applying the pruning mask at each step. The agent initially discards plain backgrounds and then removes redundant details unrelated to the target object. In the final stages, the policy eliminates residual redundancy within the object itself while preserving vital semantic features. 
This progressive filtering demonstrates the clear advantages of our multi-step MDP formulation in narrowing down task-critical information through iterative and precise decision-making. These trajectories confirm that TPRL effectively identifies and preserves semantically vital cues while progressively stripping away non-essential visual tokens.


\section{Conclusion}
This paper represents an advancement in efficient multimodal inference by introducing TPRL as a reinforcement learning framework that formulates visual token pruning as a sequential Markov Decision Process. By optimizing through task-specific rewards our method autonomously discovers adaptive pruning trajectories. Evaluations demonstrate TPRL achieves near-lossless performance while significantly reducing visual tokens across multiple benchmarks and scales. Future work will investigate extending this adaptive visual token pruning strategy to complex video world models.


%
%
\bibliographystyle{splncs04}
\bibliography{main}
\end{document}